\documentclass[runningheads]{llncs}
\usepackage{graphicx}

\usepackage{tikz}
\usepackage{comment}
\usepackage{amsmath,amssymb} 
\usepackage{color}
\usepackage{multirow} 
\usepackage[export]{adjustbox}
\usepackage{subcaption}
\usepackage{amsfonts}
\usepackage{xspace}
\usepackage{multirow}
\usepackage[toc,title,page]{appendix}
\usepackage{floatrow}
\newfloatcommand{capbtabbox}{table}[][\FBwidth]
\DeclareRobustCommand\onedot{\futurelet\@let@token\@onedot}
\def\@onedot{\ifx\@let@token.\else.\null\fi\xspace}

\makeatother
\usepackage[accsupp]{axessibility}  
\usepackage[colorlinks=true]{hyperref}
\usepackage[misc]{ifsym}

\begin{document}
\pagestyle{headings}
\mainmatter
\def\ECCVSubNumber{3043}  

\title{Simple Baselines for Image Restoration}

\titlerunning{Nonlinear Activation Free Network}
\authorrunning{Chen et al.}
\author{Liangyu Chen\thanks{Equally contribution.} \and Xiaojie Chu$^{\star}$ \and Xiangyu Zhang \and Jian Sun\\ 
}

\institute{MEGVII Technology, Beijing, CN 
\email{\{chenliangyu,chuxiaojie,zhangxiangyu,sunjian\}@megvii.com}}
\maketitle

\begin{abstract}
Although there have been significant advances in the field of image restoration recently, the system complexity of the state-of-the-art (SOTA) methods is increasing as well, which may hinder the convenient analysis and comparison of methods. 
In this paper, we propose a simple baseline that exceeds the SOTA methods and is computationally efficient. 
To further simplify the baseline, we reveal that the nonlinear activation functions, e.g. Sigmoid, ReLU, GELU, Softmax, etc. are not necessary: they could be replaced by multiplication or removed. Thus, we derive a Nonlinear Activation Free Network, namely NAFNet, from the baseline. SOTA results are achieved on various challenging benchmarks, e.g. 33.69 dB PSNR on GoPro (for image deblurring), exceeding the previous SOTA 0.38 dB with only 8.4\% of its computational costs; 40.30 dB PSNR on SIDD (for image denoising), exceeding the previous SOTA 0.28 dB with less than half of its computational costs.
The code and the pre-trained models are released at \href{https://github.com/megvii-research/NAFNet}{github.com/megvii-research/NAFNet}.
\end{abstract}

\section{Introduction}
With the development of deep learning, the performance of image restoration methods improve significantly. Deep learning based methods\cite{chen2021hinet,waqas2021multi,zamir2021restormer,wang2021uformer,cheng2021nbnet,cho2021rethinking,tu2022maxim,chu2021revisiting,mao2021deep} have achieved tremendous success. E.g. \cite{zamir2021restormer} and \cite{chu2021revisiting} achieve 40.02/33.31 dB of PSNR on SIDD\cite{SIDD_2018_CVPR}/GoPro\cite{nah2017deep} for image denoising/deblurring respectively.



Despite their good performance, these methods suffer from high system complexity. For a clear discussion, we decompose the system complexity into two parts: 
inter-block complexity and intra-block complexity.
First, the inter-block complexity, as shown in Figure~\ref{fig:compares-archs}. \cite{cho2021rethinking,mao2021deep} introduce connections between various-sized feature maps; \cite{chen2021hinet,waqas2021multi} are multi-stage networks and the latter stage refine the results of the previous stage. 
Second, the intra-block complexity, i.e. the various design choices inside the block. E.g. Multi-Dconv Head Transposed Attention Module and Gated Dconv Feed-Forward Network in \cite{zamir2021restormer} (as we shown in Figure~\ref{fig:compares-blocks}a), Swin Transformer Block in \cite{liu2021swin}, HINBlock in \cite{chen2021hinet}, and etc. It is not practical to evaluate the design choices one by one. 

Based on the above facts, a natural question arises: 
Is it possible that a network with low inter-block and low intra-block complexity can achieve SOTA performance?
To accomplish the first condition (low inter-block complexity), this paper adopts the single-stage UNet as architecture (following some SOTA methods\cite{zamir2021restormer,wang2021uformer}) and focuses on the second condition. 
To this end, we start with a plain block with the most common components, i.e. convolution, ReLU, and shortcut\cite{he2016deep}. From the plain block, we add/replace components of SOTA methods and verify how much performance gain do these components bring. 
By extensive ablation studies, we propose a simple baseline, as shown in Figure~\ref{fig:compares-blocks}c, that exceeds the SOTA methods and is computationally efficient. 
It has the potential to inspire new ideas and make their verification easier. 
The baseline, which contains GELU\cite{hendrycks2016gaussian} and Channel Attention Module\cite{hu2018squeeze} (CA), can be further simplified: we reveal that the GELU in the baseline can be regarded as a special case of the Gated Linear Unit\cite{dauphin2017language} (GLU), and from this we empirically demonstrate that it can be replaced by a simple gate, i.e. element-wise product of feature maps.
In addition, we reveal the similarity of the CA to GLU in form, and the nonlinear activation functions in CA could be removed either. In conclusion, the simple baseline could be further simplified to a nonlinear activation free network, noted as NAFNet.
\begin{figure*}[!t]
\includegraphics[width=1.0\textwidth]{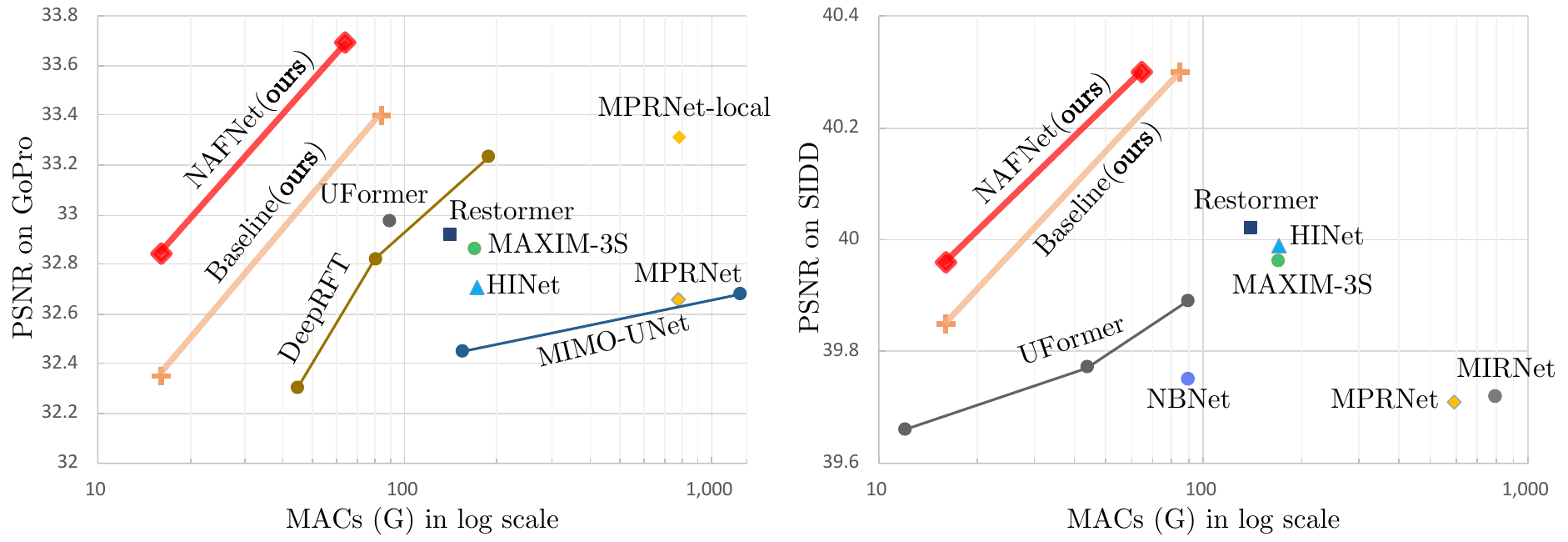}
\caption{PSNR vs. computational cost on Image Deblurring (left) and Image Denoising (right) tasks}
\label{fig:PSNR_vs_MACs}
\end{figure*}
We mainly conduct experiments on SIDD\cite{SIDD_2018_CVPR} for image denoising, and  GoPro\cite{nah2017deep} for image deblurring, following \cite{chen2021hinet,zamir2021restormer,waqas2021multi}.
The main results are shown in Figure~\ref{fig:PSNR_vs_MACs}, our proposed baseline and NAFNet achieves SOTA results while being computationally efficient: 33.40/33.69 dB on GoPro, exceed previous SOTA\cite{chu2021revisiting} 0.09/0.38 dB, respectively, with 8.4\% of its computational cost; 40.30 dB on SIDD, exceed \cite{zamir2021restormer} 0.28 dB with less than half of its computational costs. Extensive quantity and quality experiments are conducted to illustrate the effectiveness of our proposed baselines. 


The contributions of this paper are summarized as follows:
 \begin{enumerate}
\item By decomposing the SOTA methods and extracting their essential components, we form a baseline (in Figure~\ref{fig:compares-blocks}c) with lower system complexity, which can exceed the previous SOTA methods and has a lower computational cost, as shown in Figure~\ref{fig:PSNR_vs_MACs}. 
It may facilitate the researchers to inspire new ideas and evaluate them conveniently. 
 \item By revealing the connections between GELU, Channel Attention to Gated Linear Unit, we further simplify the baseline  by removing or replacing the nonlinear activation functions (e.g. Sigmoid, ReLU, and GELU), and propose a nonlinear activation free network, namely NAFNet. It can match or surpass the baseline although being simplified. To the best of our knowledge, it is the first work demonstrates that the nonlinear activation functions may not be necessary for SOTA computer vision methods. This work may have the potential to expand the design space of SOTA computer vision methods.

 \end{enumerate}




\section{Related Works}

\subsection{Image Restoration}
Image restoration tasks aim to restore a degraded image (e.g. noisy, blur) to a clean one. 
Recently, deep learning based methods\cite{chen2021hinet,waqas2021multi,zamir2021restormer,wang2021uformer,cheng2021nbnet,cho2021rethinking,tu2022maxim,chu2021revisiting,mao2021deep} achieve SOTA results on these tasks, and most of the methods could be viewed as variants of a classical solution, UNet\cite{ronneberger2015u}. It stacks blocks to a U-shaped architecture with skip-connection. 
The variants bring performance gain, as well as the system complexity, and we broadly categorized the complexity as inter-block complexity and intra-block complexity.

\subsubsection{Inter-block Complexity}
\cite{waqas2021multi,chen2021hinet} are multi-stage networks, i.e. the latter stage refine the results of the previous stage, and each stage is a U-shaped architecture. This design is based on the assumption that breaking down the difficult image restoration task into several subtasks contributes to performance. Differently, \cite{cho2021rethinking,mao2021deep} adopt the single-stage design and achieve competitive results, but they introduce complicated connections between various sized feature maps. Some methods adopt the above strategies both, e.g. \cite{tu2022maxim}. Other SOTA methods, e.g. \cite{zamir2021restormer,wang2021uformer} maintain the simple structure of single-stage UNet, yet they introduce intra-block complexity, which we will discuss next.

\subsubsection{Intra-block Complexity}
There are numerous different intra-block design schemes, we pick a few examples here. \cite{zamir2021restormer} reduces the memory and time complexity of self-attention\cite{vaswani2017attention} by channelwise attention map rather than spatialwise. Besides, gated linear units\cite{dauphin2017language} and depthwise convolution are adopted in the feed-forward network. \cite{wang2021uformer} introduces window-based multi-head self-attention, which is similar to \cite{liu2021swin}. In addition, it introduces locally-enhanced feed-forward network in its block, which adds depthwise convolution to feed-forward network to enhance the local information capture ability. Differently, we reveal that increasing system complexity is not the only way to improve performance: SOTA performance could be achieved by a simple baseline.

\subsection{Gated Linear Units}
Gated Linear Units\cite{dauphin2017language} (GLU) can be interpreted by the element-wise production of two linear transformation layers, one of which is activated with the nonlinearity. GLU or its variants has verified their effectiveness in NLP\cite{shazeer2020glu,dauphin2017language,dai2019transformer}, and there is a prosperous trend of them in computer vision\cite{tu2022maxim,zamir2021restormer,hua2022transformer,liang2022vrt}. 
In this paper, we reveal the non-trivial improvement brought by GLU. Different from \cite{shazeer2020glu}, we remove the nonlinear activation function in GLU without performance degradation.
Furthermore, based on the fact that the nonlinear activation free GLU contains nonlinearity itself (as the product of two linear transformations raises nonlinearity), our baseline could be simplified by replacing the nonlinear activation functions with the multiplication of two feature maps. To the best of our knowledge, it is the first computer vision model achieves SOTA performance without nonlinear activation functions. 


\begin{figure*}[!t]
\includegraphics[width=1.0\textwidth]{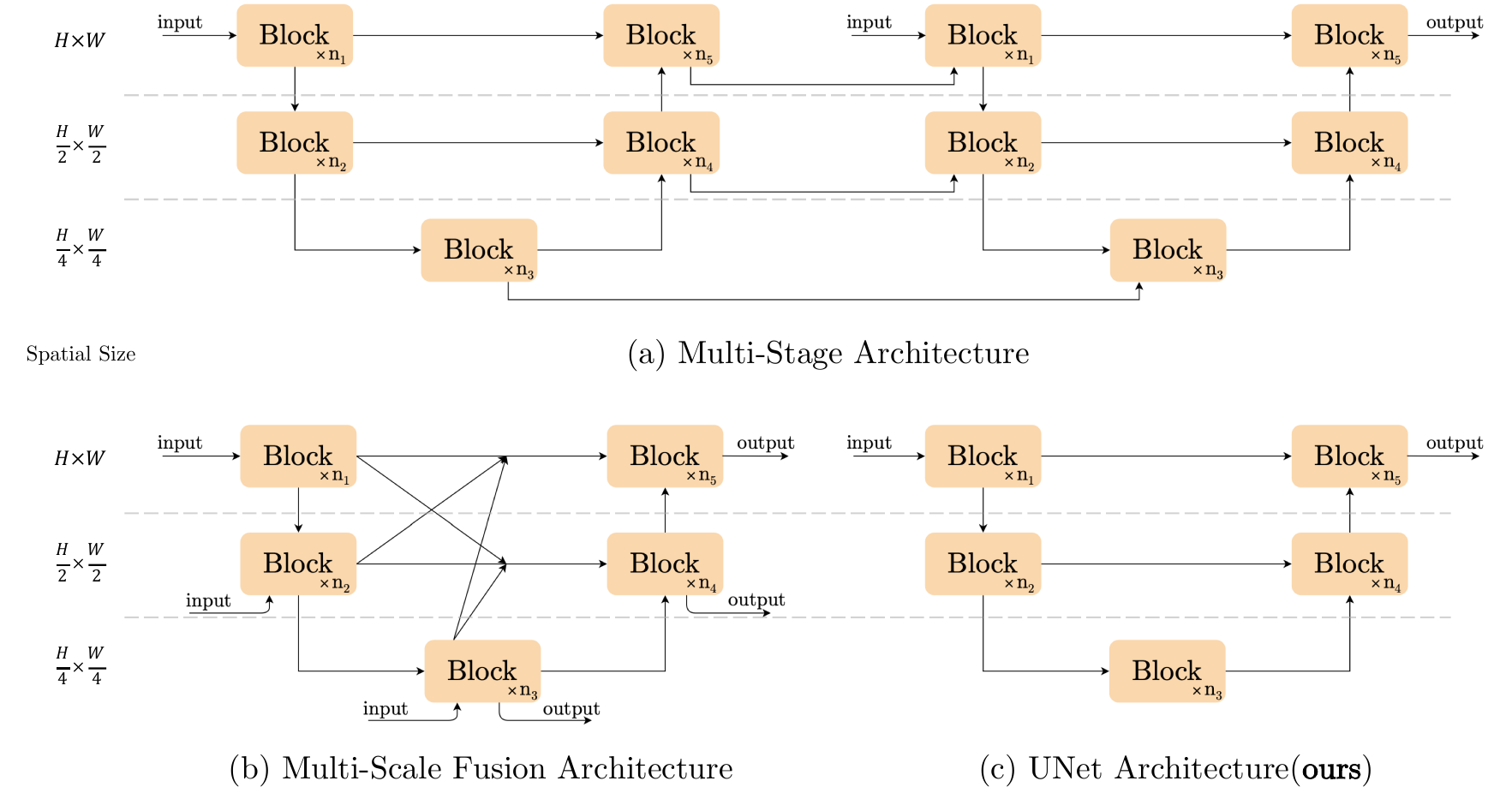}
\caption{Comparison of architectures of image restoration models. Dashes to distinguish features of different sizes. (a) The multi-stage architecture\cite{chen2021hinet,waqas2021multi} stacks UNet architecture serially. (b) The multi-scale fusion architecture\cite{mao2021deep,cho2021rethinking} fusions the features in different scales. (c)UNet architecture, which is adopted by some SOTA methods\cite{zamir2021restormer,wang2021uformer}. We use it as our architecture. Some details have been deliberately omitted for simplicity, e.g. downsample/upsample layers, feature fusion modules, input/output shortcut, and etc. }
\label{fig:compares-archs}
\end{figure*}

\section{Build A Simple Baseline}
In this section, we build a simple baseline for image restoration tasks from scratch.  To keep the structure simple, our principle is not to add entities if they are not necessary. The necessity is verified by empirical evaluation of restoration tasks.
We mainly conduct experiments with the model size around 16 GMACs following HINet Simple\cite{chen2021hinet}, and the MACs are estimated by an input with the spatial size of $256\times 256$. The results of models with different capacities are in the experimental section. We mainly validate the results (PSNR) on two popular datasets for denoising (i.e. SIDD\cite{SIDD_2018_CVPR}) and deblurring (i.e. GoPro\cite{nah2017deep} dataset), based on the fact that those tasks are fundamental in low-level vision. The design choices are discussed in the following subsections.

\subsection {Architecture}
To reduce the inter-block complexity, we adopt the classic single-stage U-shaped architecture with skip-connections, as shown in Figure~\ref{fig:compares-archs}c, following \cite{zamir2021restormer,wang2021uformer}. We believe the architecture will not be a barrier to performance. The experimental results confirmed our conjecture, in Table~\ref{tab:SOTAs-on-SIDD}, ~\ref{tab:SOTAs-on-GoPro} and Figure~\ref{fig:PSNR_vs_MACs}.

\subsection{A Plain Block}
\begin{figure*}[!t]
\includegraphics[width=1.0\textwidth]{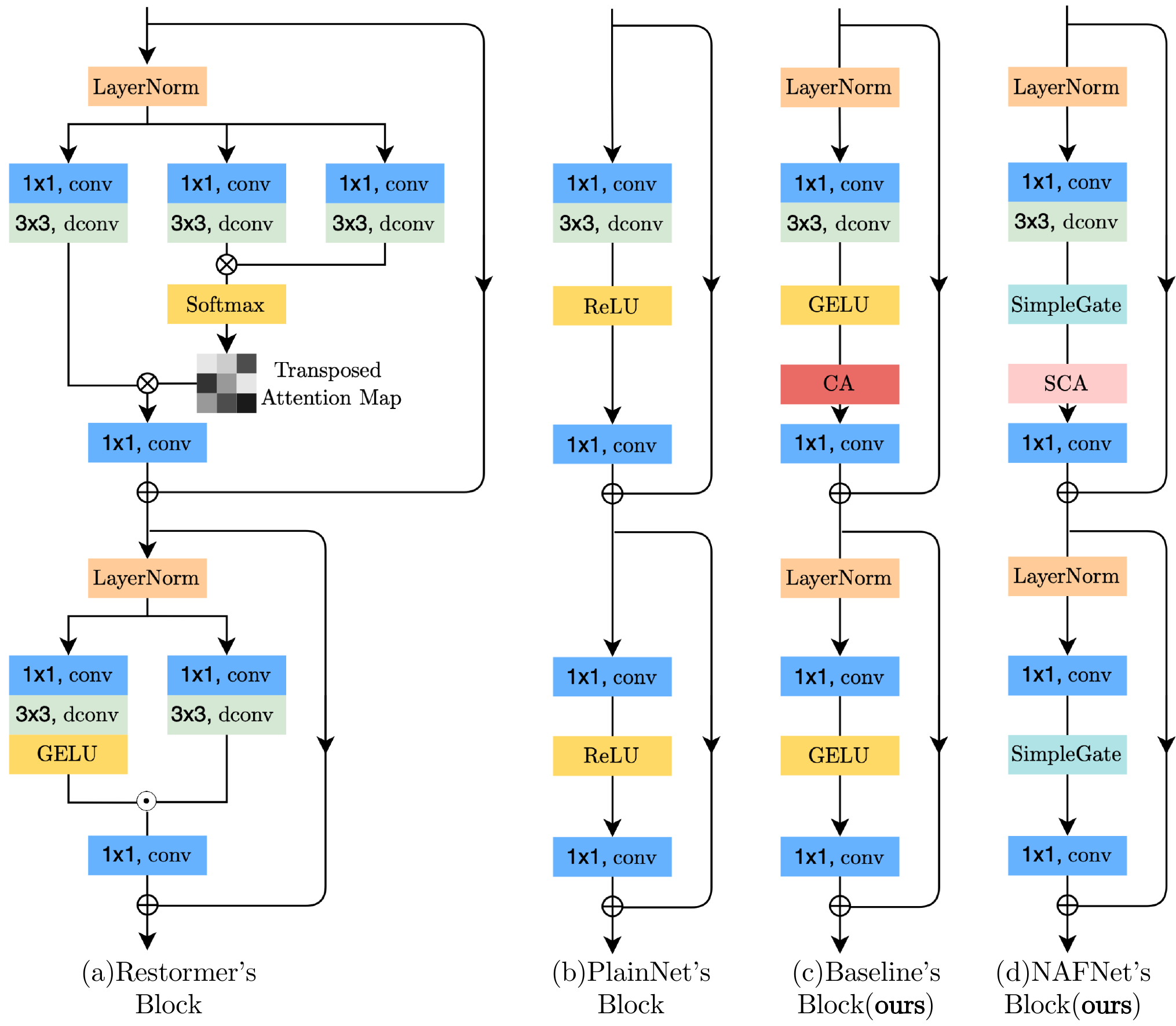}
\caption{Intra-block structure comparison. $\otimes$:matrix multiplication, $\odot$/$\oplus$:element-wise multiplication/addition. dconv: Depthwise convolution. Nonlinear activation functions are represented by yellow boxes. (a) Restormer's block\cite{zamir2021restormer}, some details are omitted for simplicity, e.g. reshaping the feature maps. (b) PlainNet's block, which contains the most common components. (c) Our proposed baseline. Compares to (b), Channel Attention (CA) and LayerNorm are adopted. Besides, ReLU is replaced by GELU. (d) Our proposed Nonlinear Activation Free Network's block. It replaces CA/GELU with Simplified Channel Attention(SCA) and SimpleGate respectively. The details of these components are shown in Fig~\ref{fig:OurModules}}
\label{fig:compares-blocks}
\end{figure*}

Neural Networks are stacked by blocks. We have determined how to stack blocks in the above (i.e. stacked in a UNet architecture), but how to design the internal structure of the block is still a problem. We start from a plain block with the most common components, i.e. convolution, ReLU, and shortcut\cite{he2016deep}, and the arrangement of these components follows \cite{han2021demystifying,liu2021swin}, as shown in Figure~\ref{fig:compares-blocks}b. We will note it as PlainNet for simplicity. Using a convolution network instead of a transformer is based on the following considerations. First, although transformers show good performance in computer vision, some works\cite{han2021demystifying,liu2022convnet} claim that they may not be necessary for achieving SOTA results. Second, depthwise convolution is simpler than the self-attention\cite{vaswani2017attention} mechanism. Third, this paper is not intended to discuss the advantages and disadvantages of transformers and convolutional neural networks, but just to provide a simple baseline. The discussion of the attention mechanism is proposed in the subsequent subsection. 

\subsection{Normalization}
Normalization is widely adopted in high-level computer vision tasks, and there is also a popular trend in low-level vision. 
Although \cite{nah2017deep} abandoned Batch Normalization\cite{ioffe2015batch} as the small batch size may bring the unstable statistics\cite{yan2020towards}, \cite{chen2021hinet} re-introduce the Instance Normalization\cite{ulyanov2016instance} and avoids the small batch size issue. However, \cite{chen2021hinet} shows that adding instance normalization does not always bring performance gains and requires manual tuning. Differently, under the prosperity of transformers, Layer Normalization\cite{ba2016layer} is used by more and more methods, including SOTA methods\cite{tu2022maxim,zamir2021restormer,wang2021uformer,liu2022convnet,liu2021swin}. Based on these facts we conjecture Layer Normalization may be crucial to SOTA restorers, thus we add Layer Normalization to the plain block described above. This change can make training smooth, even with a $10\times$ increase in learning rate. The larger learning rate brings significant performance gain: +0.44 dB (39.29 dB to 39.73 dB) on SIDD\cite{SIDD_2018_CVPR}, +3.39 dB (28.51 dB to 31.90 dB) on GoPro\cite{nah2017deep} dataset. To sum up, we add Layer Normalization to the plain block as it can stabilize the training process.


\subsection{Activation}
The activation function in the plain block, Rectified Linear Unit\cite{nair2010rectified} (ReLU), is extensively used in computer vision. However, there is a tendency to replace ReLU with GELU\cite{hendrycks2016gaussian} in SOTA methods\cite{liu2022convnet,zamir2021restormer,tu2022maxim,liu2021swin,dosovitskiy2020image}. 
This replacement is implemented in our model either. The performance stays comparable on SIDD (from 39.73 dB to 39.71 dB) which is consistent with the conclusion of \cite{liu2022convnet}, yet it brings 0.21 dB performance gain (31.90 dB to 32.11 dB) on GoPro. In short, we replace ReLU with GELU in the plain block, because it keeps the performance of image denoising while bringing non-trivial gain on image deblurring.


\subsection{Attention}
Considering the recent popularity of the transformer in computer vision, its attention mechanism is an unavoidable topic in the design of the internal structure of the block. 
There are many variants of attention mechanisms, and we discuss only a few of them here.
The vanilla self-attention mechanism\cite{vaswani2017attention}, which is adopted by \cite{dosovitskiy2020image,chen2021pre}, generate the target feature by the linear combination of all features which are weighted by the similarity between them. Therefore, each feature contains global information, while it suffers from the quadratic computational complexity with the size of the feature map.
Some image restoration tasks process data at high resolution which makes the vanilla self-attention not practical.
Alternatively, \cite{liu2021swin,liang2021swinir,wang2021uformer} apply self-attention only in a fix-sized local window to alleviate the issue of increased computation. While it lacks global information. We do not take the window-based attention, as the local information could be well captured by the depthwise convolution \cite{han2021demystifying,liu2022convnet} in the plain block. 

Differently, \cite{zamir2021restormer} modifies the spatial-wise attention to channel-wise, avoids the computation issue while maintaining global information in each feature. It could be seen as a special variant of channel attention \cite{hu2018squeeze}. Inspired by \cite{zamir2021restormer}, we realize the vanilla channel attention meets the requirements: computational efficiency and brings global information to the feature map. In addition, the effectiveness of channel attention has been verified in the image restoration task\cite{waqas2021multi,chu2021revisiting}, thus we add the channel attention to the plain block. It brings 0.14 dB on SIDD\cite{SIDD_2018_CVPR} (39.71 dB to 39.85 dB), 0.24 dB on GoPro\cite{nah2017deep} dataset (32.11 dB to 32.35 dB). 

\subsection{Summary}
So far, we build a simple baseline from scratch, as we shown in Table~\ref{tab:ablation-build_baseline}. The architecture and the block are shown in Figure~\ref{fig:compares-archs}c and Figure~\ref{fig:compares-blocks}c, respectively. 
Each component in the baseline is trivial, e.g. Layer Normalization, Convolution, GELU, and Channel Attention. But the combination of these trivial components leads to a strong baseline: it can surpass the previous SOTA results on SIDD and GoPro dataset with only a fraction of computation costs, as we shown in Figure~\ref{fig:PSNR_vs_MACs} and Table~\ref{tab:SOTAs-on-SIDD},\ref{tab:SOTAs-on-GoPro}.
We believe the simple baseline could facilitate the researchers to evaluate their ideas.

\section{Nonlinear Activation Free Network}
The baseline described above is simple and competitive, but is it possible to further improve performance while ensuring simplicity?
Can it be simpler without performance loss? 
We try to answer these questions by looking for commonalities from some SOTA methods\cite{tu2022maxim,zamir2021restormer,liang2022vrt,hua2022transformer}. We find that in these methods, Gated Linear Units\cite{dauphin2017language}(GLU) are adopted. It implies that GLU might be promising. We will discuss it next.
\subsubsection{Gated Linear Units}
\begin{figure*}[!t]
\includegraphics[width=1.0\textwidth]{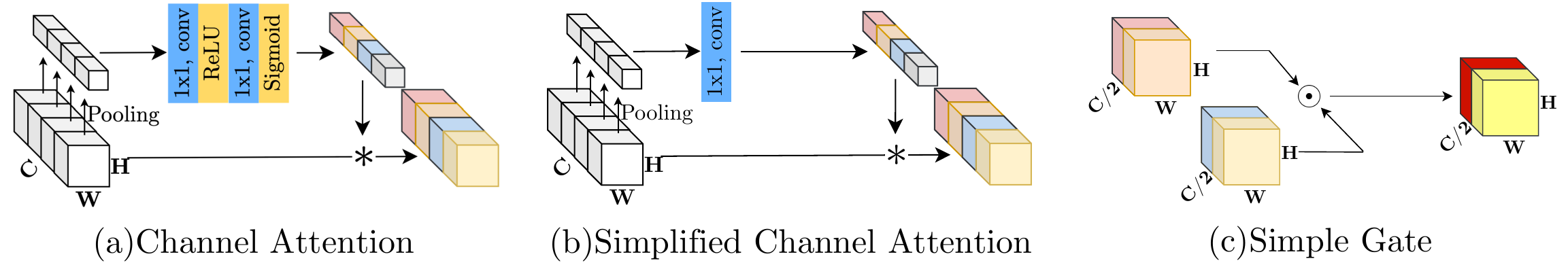}
\caption{ Illustration of (a) Channel Attention\cite{hu2018squeeze} (CA), (b) Simplified Channel Attention (SCA), and (c) Simple Gate (SG). $\odot$/$*$: element-wise/channel-wise multiplication
}
\label{fig:OurModules}
\end{figure*}

The gated linear units could be formulated as:
\begin{equation}\label{eqn:gate}
Gate(\mathbf{X}, f, g, \sigma) = f(\mathbf{X})\odot\sigma(g(\mathbf{X})),
\end{equation}
where $\mathbf{X}$ represents the feature map, $f$ and $g$ are linear transformers, $\sigma$ is a non-linear activation function, e.g. Sigmoid, and $\odot$ indicates element-wise multiplication. As discussed above, adding GLU to our baseline may improve the performance yet the intra-block complexity is increasing as well. This is not what we expected. To address this, we revisit the activation function in the baseline, i.e. GELU\cite{hendrycks2016gaussian}:
\begin{equation}\label{eqn:gelu}
GELU(x) =x \Phi(x),
 \end{equation}
 where $\Phi$ indicates the cumulative distribution function of the standard normal distribution. And based on~\cite{hendrycks2016gaussian}, GELU could be approximated and implemented by:
 \begin{equation}\label{eqn:gelu-app}
 0.5x(1+tanh[\sqrt{2/\pi}(x+0.044715x^3)]).
 \end{equation}
From Eqn.~\ref{eqn:gate} and Eqn.~\ref{eqn:gelu}, it can be noticed that GELU is a special case of GLU, i.e. $f$, $g$ are identity functions and take $\sigma$ as $\Phi$. 
Through the similarity, we conjecture from another perspective that GLU may be regarded as a generalization of activation functions, and it might be able to replace the nonlinear activation functions.
Further, we note that the GLU itself contains nonlinearity and does \emph{not} depend on $\sigma$: even if the $\sigma$ is removed, $Gate(\mathbf{X})=f(\mathbf{X})\odot g(\mathbf{X})$ contains nonlinearity. Based on these, we propose a simple GLU variant: directly divide the feature map into two parts in the channel dimension and multiply them, as we shown in Figure~\ref{fig:OurModules}c, noted as SimpleGate. Compared to the complicated implementation of GELU in Eqn.\ref{eqn:gelu-app}, our SimpleGate could be implemented by an element-wise multiplication, that's all:
\begin{equation}\label{eqn:simple-gate}
SimpleGate(\mathbf{X},\mathbf{Y}) = \mathbf{X\odot Y},
\end{equation}
where $\mathbf{X}$ and $\mathbf{Y}$ are feature maps of the same size.

By replacing GELU in the baseline to the proposed SimpleGate, the performance of image denoising (on SIDD\cite{SIDD_2018_CVPR}) and image deblurring (on GoPro\cite{nah2017deep} dataset) boost 0.08 dB (39.85 dB to 39.93 dB) and 0.41 dB (32.35 dB to 32.76 dB) respectively. The results demonstrate that GELU could be replaced by our proposed SimpleGate. At this point, only a few types of nonlinear activations left in the network: Sigmoid and ReLU in the channel attention module\cite{hu2018squeeze}, and we will discuss the simplifications of it next.

\subsubsection{Simplified Channel Attention}
In Section 3, we adopt the channel attention\cite{hu2018squeeze} into our block as it captures the global information and it is computationally efficient. It is illustrated in Figure~\ref{fig:OurModules}a: it squeezes the spatial information into channels first and then a multilayer perceptual applies to it to calculate the channel attention, which will be used to weight the feature map. It could be represented as:
\begin{equation}\label{eqn:channel-attention}
CA(\mathbf{X}) = \mathbf{X} * \sigma(W_2 max(0, W_1 pool(\mathbf{X}))),
\end{equation} 
where $\mathbf{X}$ represents the feature map, $pool$ indicates the global average pooling operation which aggregates the spatial information into channels. $\sigma$ is a nonlinear activation function, Sigmoid, $W_1, W_2$ are fully-connected layers and ReLU is adopted between two fully-connected layers. Last, $*$ is a channelwise product operation. If we regard the channel-attention calculation as a function, noted as $\Psi$ with input $\mathbf{X}$ , Eqn.~\ref{eqn:channel-attention} could be re-writed as:
\begin{equation}\label{eqn:channel-attention-func}
CA(\mathbf{X}) = \mathbf{X} * \Psi(\mathbf{X}).
\end{equation}
It can be noticed that Eqn.~\ref{eqn:channel-attention-func} is very similar to Eqn.~\ref{eqn:gate}. 
This inspires us to consider channel attention as a special case of GLU, which can be simplified like GLU in the previous subsection.
By retaining the two most important roles of channel attention, that is, aggregating global information and channel information interaction, we propose the Simplified Channel Attention:
\begin{equation}\label{eqn:simplified-channel-attention}
SCA(\mathbf{X}) = \mathbf{X} * W pool(\mathbf{X}).
\end{equation}
The notations follows Eqn.~\ref{eqn:channel-attention}. Apparently, Simplified Channel Attention (Eqn.~\ref{eqn:simplified-channel-attention}) is simpler than the original one (Eqn.~\ref{eqn:channel-attention}), as shown in Figure~\ref{fig:OurModules}a and Figure~\ref{fig:OurModules}b.  
Although it is simpler, there is no loss of performance: +0.03 dB (39.93 dB to 39.96 dB) on SIDD and +0.09 dB  (32.76 dB to 32.85 dB) on GoPro. 

\subsubsection{Summary}
Starting from the baseline proposed in Section 3, we further simplify it
by replacing the GELU with SimpleGate and Channel Attention to Simplified Channel Attention, without loss of performance.
We emphasize that after the simplification, there are \emph{no} nonlinear activation functions (e.g. ReLU, GELU, Sigmoid, etc.) in the network. So we call this baseline Nonlinear Activation Free Network, namely NAFNet.
It can match or surpass the baseline although without nonlinear activation functions, as we shown in Figure~\ref{fig:PSNR_vs_MACs} and Table~\ref{tab:SOTAs-on-SIDD},\ref{tab:SOTAs-on-GoPro}.
We can now answer the questions in the begining of this section by yes, because of the simplicity and effectiveness of NAFNet.

\begin{figure*}[!t]
\centering
\scalebox{0.98}{
\tiny
\begin{tabular}{ccccccc }
\includegraphics[width=0.14\textwidth]{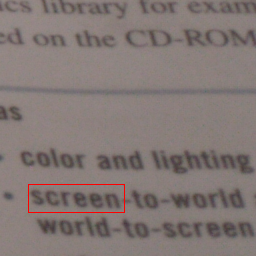} &
\includegraphics[width=0.14\textwidth]{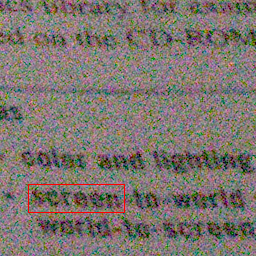} &
\includegraphics[width=0.14\textwidth]{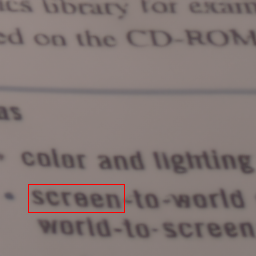} &
\includegraphics[width=0.14\textwidth]{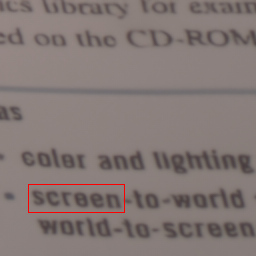} &
\includegraphics[width=0.14\textwidth]{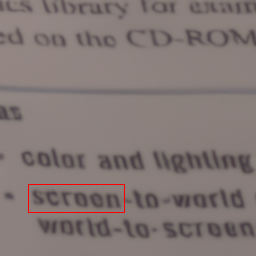} &
\includegraphics[width=0.14\textwidth]{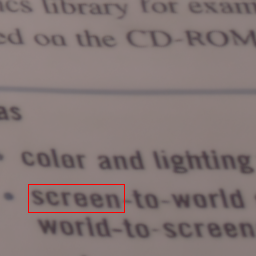} &
\includegraphics[width=0.14\textwidth]{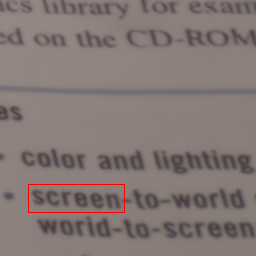} 
\\
\includegraphics[width=0.14\textwidth]{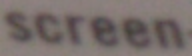} &
\includegraphics[width=0.14\textwidth]{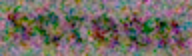} &
\includegraphics[width=0.14\textwidth]{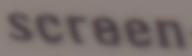} &
\includegraphics[width=0.14\textwidth]{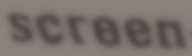} &
\includegraphics[width=0.14\textwidth]{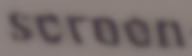} &
\includegraphics[width=0.14\textwidth]{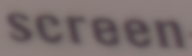} &
\includegraphics[width=0.14\textwidth]{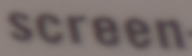} 
\\
PSNR & 19.01 dB & 35.21 dB& ~35.01 dB & ~34.96 dB & ~35.97 dB& ~35.77 dB \\
Reference & ~Noisy & ~HINet\cite{chen2021hinet} & ~Restormer\cite{zamir2021restormer} & ~MPRNet\cite{waqas2021multi} & Baseline(\textbf{ours}) & {NAFNet(\textbf{ours})}\\
\end{tabular}

}
\caption{Qualitative comparison of image denoising methods on SIDD\cite{SIDD_2018_CVPR}
}
\label{fig:visual-sidd}
\end{figure*}

\section{Experiments}
In this section, we analyze the effect of the design choices of NAFNet described in previous sections in detail. Next, we apply our proposed NAFNet to various image restoration applications, including RGB image denoising, image deblurring, raw image denoising, and image deblurring with JPEG artifacts.
\subsection{Ablations}
The ablation studys are conducted on image denoising (SIDD\cite{SIDD_2018_CVPR}) and deblurring (GoPro\cite{nah2017deep}) tasks. 
 We follow experiments setting of \cite{chen2021hinet} if not specified, e.g. 16 GMACs of computational budget, gradient clip, and PSNR loss.
We train models with Adam\cite{kingma2014adam} optimizer ($\beta_1=0.9, \beta_2=0.9$, weight decay 0) for total 200K iterations with the initial learning rate $1e^{-3}$ gradually reduced to $1e^{-6}$ with the cosine annealing schedule\cite{loshchilov2016sgdr}. The training patch size is $256\times 256$ and batch size is 32. Training by patches and testing by the full image raises performance degradation\cite{chu2021revisiting}, we solve it by adopting TLC\cite{chu2021revisiting} following MPRNet-local\cite{chu2021revisiting}. 
The effectiveness of TLC on GoPro\footnote{SIDD test on $256\times256$ patches avoid the inconsistent issue.} is shown in Tab~\ref{tab:ablation-tlc}. We mainly compare TLC with ``test by patches'' strategy, which is adopted by \cite{chen2021hinet}, \cite{mao2021deep}, and etc. It brings performance gains and avoids the artifacts brought by patches.
Moreover,  we apply skip-init\cite{de2020batch} to stabilize training following \cite{liu2022convnet}. The default width and number of blocks are 32 and 36, respectively. We adjust the width to keep the computational budget hold if the number of blocks changed. 
We report Peak Signal to Noise Ratio (PSNR) and Structural SIMilarity (SSIM) in our experiments. The speed/memory/computational complexity evaluation is conducted with an input size of $256\times 256$, on an NVIDIA 2080Ti GPU. 


\subsubsection{From PlainNet to the simple baseline:} PlainNet is defined in Section 3, and its block is illustrated in Figure~\ref{fig:compares-blocks}b. We find that the training of PlainNet is unstable under the default settings. As an alternative, we reduce the learning rate (lr) by a factor of 10 to make the model trainable. This issue is solved by introducing Layer Normalization (LN): the learning rate can be increased from $1e^{-4}$ to $1e^{-3}$ with a more stable training process. In PSNR, LN brings 0.46 dB and 3.39 dB on SIDD and GoPro respectively. Besides, GELU and Channel Attention (CA) also demonstrated their effectiveness in Table~\ref{tab:ablation-build_baseline}. 


\begin{table}
\begin{center}
\caption{Build a simple baseline from PlainNet. The effectiveness of Layer Normalization (LN), GELU, and Channel Attention (CA) have been verified. $*$ indicates that the training is unstable due to the large learning rate (lr)}
\label{tab:ablation-build_baseline}
\begin{tabular}{c|c|c|c|c|cc|cc}
\hline
 & \multirow{2}{*}{\ \ \ lr\ \ \ } & \multirow{2}{*}{\ \ \ \ \ LN\ \ \ \ \ } & \multirow{2}{*}{\ ReLU$\rightarrow$GELU\ }& \multirow{2}{*}{\ \ \ \ \ CA\ \ \ \ \ } &\multicolumn{2}{|c|}{SIDD} & \multicolumn{2}{|c}{GoPro} \\
  & & & & &PSNR & SSIM & PSNR & SSIM\\
  \hline
 PlainNet & $1e^{-4}$ & & & & 39.29 & 0.956 & 28.51 & 0.907 \\
 PlainNet$^*$ & $1e^{-3}$  & & & & - & - & - & - \\
          & $1e^{-3}$ & \checkmark & & & 39.73 & 0.959 & 31.90 & 0.952 \\
          & $1e^{-3}$ & \checkmark & \checkmark & & 39.71 & 0.958 & 32.11 & 0.954 \\
 Baseline & $1e^{-3}$ & \checkmark & \checkmark & \checkmark & 39.85 & 0.959 & 32.35 & 0.956 \\
 \hline
\end{tabular}
\end{center}
\end{table}
\begin{table}
\begin{center}
\caption{NAFNet is derived from the simplification of baseline, i.e. replacing GELU to SimpleGate (SG), and replacing Channel Attention (CA) to Simplified Channel Attention (SCA). }
\label{tab:ablation-baseline_to_nafnet}
\begin{tabular}{c|c|c|cc|cc|c}
\hline
 & \multirow{2}{*}{\ GELU$\rightarrow$SG\ } & \multirow{2}{*}{\ CA$\rightarrow$SCA\ } &  \multicolumn{2}{|c|}{SIDD} & \multicolumn{2}{|c|}{GoPro} & \multirow{2}{*}{speedup}\\
  & & & PSNR & SSIM & PSNR & SSIM & \\
  \hline
Baseline\  &  &  & 39.85 & 0.959 & 32.35 & 0.956 & $1.00\times$ \\
 & \checkmark &  & 39.93 & 0.960 & 32.76 & 0.960  &$0.98\times$ \\
 &  & \checkmark & 39.95 & 0.960 & 32.54 & 0.958 &$1.11\times$ \\
 NAFNet\  & \checkmark & \checkmark & 39.96 & 0.960 & 32.85 & 0.960 & $1.09\times$\\
 \hline
\end{tabular}
\end{center}
\end{table}

\subsubsection{From the simple baseline to NAFNet:} As described in Section 3, NAFNet can be obtained by simplifying the baseline. 
In Tab~\ref{tab:ablation-baseline_to_nafnet}, we show that there is no performance penalty for this simplification. Instead, the PSNR boosts 0.11 dB and 0.50 dB in SIDD and GoPro respectively.
The computational complexity is consistent for a fair comparison, and details in the supplementary material.  The speedup of modifications compared to Baseline is provided. In addition, no significant extra memory consumption compares to Baseline in inference.

\subsubsection{Number of blocks:} 
We verify the effect of the number of blocks on NAFNet in Table~\ref{tab:ablation-num_of_blk}. We mainly consider the latency at spatial size $720\times1280$, as this is the size of the entire GoPro image.
In the process of increasing the number of blocks to 36, the performance of the model has been greatly improved, and the latency has not increased significantly (+14.5\% compares to 9 blocks). When the number of blocks further increases to 72, the performance improvement of the model is not obvious, but the latency increases significantly (+30.0\% compares to 36 blocks). Because 36 blocks can achieve a better performance/latency balance, we use it as the default option.

\begin{table}
\begin{center}
\caption{The effect of the number of blocks. The width is adjusted to keep the computational budget hold. Latency-256 and Latency-720 is based on the input size $256\times256$ and $720\times1280$ respectively, in milliseconds}
\label{tab:ablation-num_of_blk}
\begin{tabular}{c|c|cc|cc|c|c}
\hline
 & \multirow{2}{*}{\ \# of blocks\ } &  \multicolumn{2}{|c|}{SIDD} & \multicolumn{2}{|c|}{GoPro} & \multirow{2}{*}{\ Latency-256\ } &\multirow{2}{*}{\ Latency-720\ }\\
  & & PSNR & SSIM & PSNR & SSIM &&\\
  \hline
\multirow{4}{*}{NAFNet} & 9 & 39.78 & 0.959 & 31.79 & 0.951 & 11.8 &154.7  \\
 & 18 & 39.90 & 0.960 & 32.64 & 0.951 & 19.9 &151.7  \\
 & 36  & 39.96 & 0.960 & 32.85 & 0.959 &39.1 &177.1  \\
 & 72 & 39.95 & 0.960 & 32.88 & 0.961 &73.8 &230.1  \\
 \hline
\end{tabular}
\end{center}
\end{table}
\begin{table}[ht]
\RawFloats
\scriptsize
\setlength{\tabcolsep}{3pt}
    \parbox{.4\linewidth}{
    \centering
\caption{Effectiveness of TLC\cite{chu2021revisiting} on GoPro\cite{nah2017deep}
}
 \label{tab:ablation-tlc}
    \begin{tabular}{lcccc}

\hline
 &{patches?} & {TLC?}  & {PSNR} & {SSIM} \\ 
\hline
\multirow{3}{*}{NAFNet}& & & 33.08 &0.963\\
& $\checkmark$& & 33.65 &0.966\\
& & $\checkmark$& 33.69 &0.967\\
\hline
& & & &\\
& & & &\\
& & & &\\
\end{tabular}}
\hfill    
\parbox{.58\linewidth}{
\centering
\caption{Variants of $\sigma$ in $SimpleGate(\mathbf{X},\mathbf{Y})=\mathbf{X} \odot \sigma(\mathbf{Y})$}
 \label{tab:ablation-variants_sigma}

\begin{tabular}{c|cc|cc}
\hline
  \multirow{2}{*}{$\sigma$} &  \multicolumn{2}{|c|}{SIDD} & \multicolumn{2}{|c}{GoPro} \\
   & PSNR & SSIM & PSNR & SSIM \\
  \hline
 Identity(\textbf{ours}) & 39.96 & 0.960 & 32.85 & 0.960\\
  ReLU & 39.98 & 0.960 & 32.59 & 0.958  \\
  GELU  & 39.97 & 0.960 & 32.72 & 0.959   \\
  Sigmoid & 39.99 & 0.960 & 32.50 & 0.958   \\
  SiLU & 39.96 & 0.960 & 32.74 & 0.960   \\
 \hline
\end{tabular}
}
\end{table}

\subsubsection{Variants of $\sigma$ in SimpleGate:} Vanilla gated linear unit (GLU) contains a nonlinear activation function $\sigma$ as formulated in Eqn.~\ref{eqn:gate}. Our proposed SimpleGate, as shown in Eqn.~\ref{eqn:simple-gate} and Figure~\ref{fig:OurModules}c removes it. In other words, $\sigma$ in SimpleGate is set as an identity function. We variants the $\sigma$ from the identity function to different nonlinear activation functions in Table~\ref{tab:ablation-variants_sigma} to judge the importance of nonlinearity in $\sigma$. PSNR on SIDD is basically unaffected (fluctuates from 39.96 dB to 39.99 dB), while PSNR on GoPro drops significantly (-0.11 dB to -0.35 dB), which indicates that in NAFNet, the $\sigma$ in SimpleGate may not be necessary.
\subsection{Applications}
We apply NAFNet to various image restoration tasks, follow the training settings of ablation study if not specified, except that it is enlarged by increasing the width from 32 to 64. Besides, batch size and total training iterations are 64 and 400K respectively, following \cite{chen2021hinet}. Random crop augmentation is applied.  We report the mean of three experimental results. The baseline is enlarged to achieve better results, details in the appendix.
\subsubsection{RGB Image Denoising} We compare the RGB Image Denoising results with other SOTA methods on SIDD, show in Table~\ref{tab:SOTAs-on-SIDD}. Baseline and its simplified version NAFNet, exceed the previous best result Restormer 0.28 dB with only a fraction of its computational cost, as shown in Figure 1. The qualitative results are shown in Figure~\ref{fig:visual-sidd}. Our proposed baselines can restore more fine details compared to other methods. Moreover, we achieve SOTA result (40.15 dB) on the \href{https://www.eecs.yorku.ca/~kamel/sidd/benchmark.php}{online benchmark}, exceed previous top-ranked methods 0.23 dB.

\begin{figure*}[!t]
\begin{center}
\scalebox{0.99}{
\begin{tabular}[b]{c@{ } c@{ }  c@{ } c@{ } }
\includegraphics[width=.24\textwidth,valign=t]{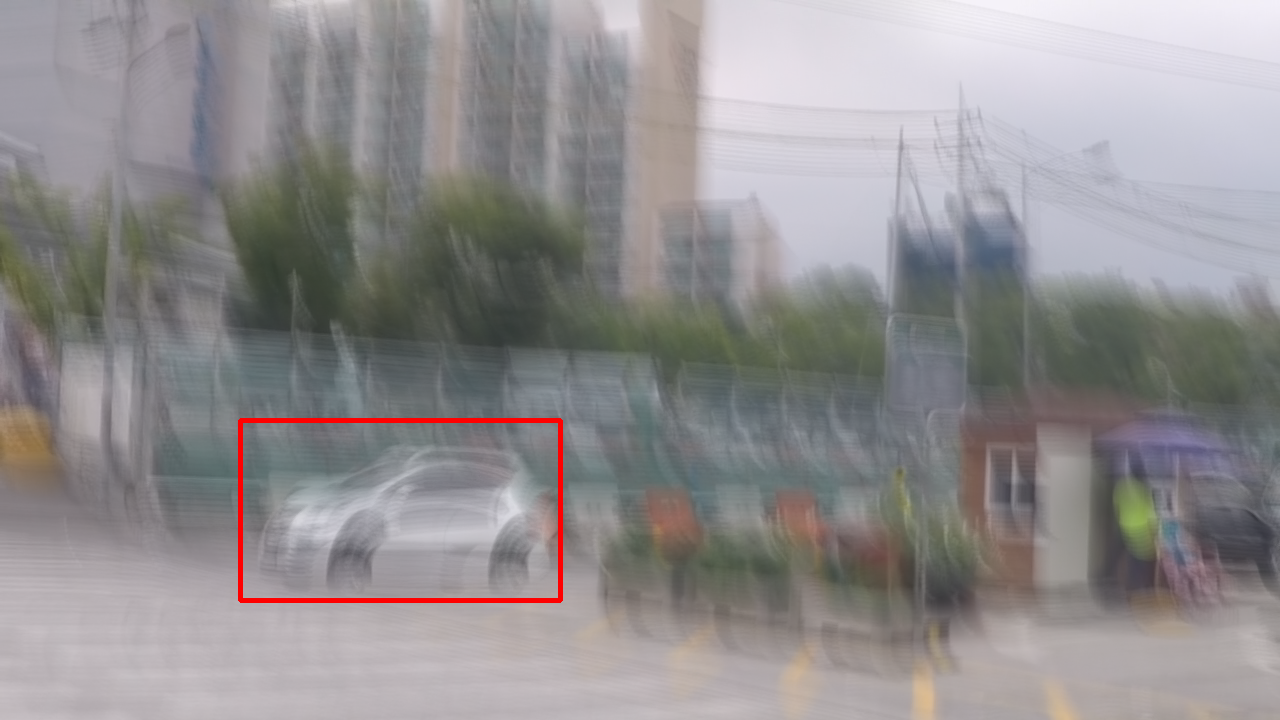}&   
    \includegraphics[width=.24\textwidth,valign=t]{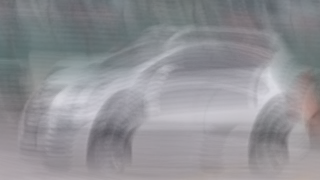} &
    \includegraphics[width=.24\textwidth,valign=t]{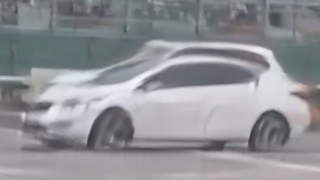} &
    \includegraphics[width=.24\textwidth,valign=t]{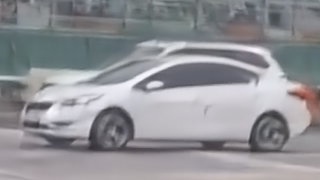} 
\\
\small~23.21 dB &  \small~23.21  &\small~29.68 dB &\small~31.58 dB   \\
   \small~Blurry Image   & \small~Blurry  & \small~Restormer~\cite{zamir2021restormer} & Baseline(\textbf{ours})  \\
\includegraphics[width=.24\textwidth,valign=t]{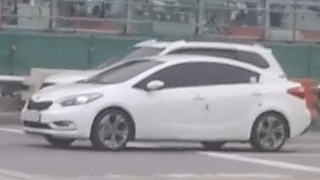}&   
    \includegraphics[width=.24\textwidth,valign=t]{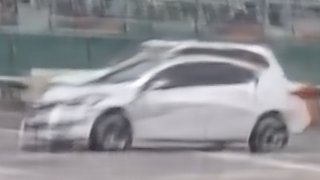} &
    \includegraphics[width=.24\textwidth,valign=t]{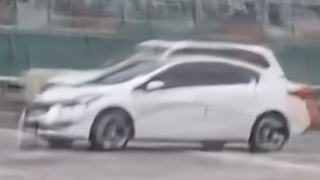} &
    \includegraphics[width=.24\textwidth,valign=t]{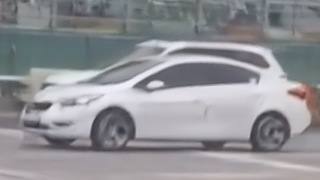}
\\
 \small~Reference  &\small~28.89 dB  &  \small~30.35 dB & \small~32.54 dB   \\
   \small~PSNR  & \small~MPRNet~\cite{waqas2021multi} & \small~MPRNet-local~\cite{chu2021revisiting} & \small~NAFNet(\textbf{ours})  \\
\end{tabular}}
\end{center}
\caption{Qualitative comparison of image deblurring methods on GoPro\cite{nah2017deep}}
\label{fig:deblur}
\end{figure*}

\begin{table}
\scriptsize
\begin{center}
\caption{Image Denoising Results on SIDD\cite{SIDD_2018_CVPR}}
\label{tab:SOTAs-on-SIDD}
\begin{tabular}{c|ccccccc|cc}
\hline
\multirow{2}{*}{Method} & MPRNet & MIRNet & NBNet & UFormer &  MAXIM & HINet &Restormer & Baseline & NAFNet \\
& \cite{waqas2021multi} & \cite{zamir2020learning} & \cite{cheng2021nbnet} & \cite{wang2021uformer} & \cite{tu2022maxim} & \cite{chen2021hinet} & \cite{zamir2021restormer} & \textbf{ours} & \textbf{ours} \\
\hline
PSNR & 39.71 & 39.72 & 39.75 & 39.89 & 39.96 & 39.99 & 40.02 & 40.30 & 40.30 \\
SSIM & 0.958 & 0.959 & 0.959 & 0.960 & 0.960 & 0.958 & 0.960 & 0.962 & 0.962 \\
\hline
MACs(G) & 588 & 786 & 88.8 & 89.5& 169.5& 170.7& 140&84& 65 \\
\hline
\end{tabular}
\end{center}
\end{table}
\vspace{-0.5cm}

\begin{table}
\scriptsize
\begin{center}
\caption{Image Deblurring Results on GoPro\cite{nah2017deep}}
\label{tab:SOTAs-on-GoPro}
\begin{tabular}{c|ccccccc|cc}
\hline
\multirow{2}{*}{Method} & MIMO-UNet & HINet & MAXIM & Restormer &  UFormer & DeepRFT & MPRNet & Baseline & NAFNet \\
                        & \cite{cho2021rethinking} & \cite{chen2021hinet} & \cite{tu2022maxim} & \cite{zamir2021restormer} & \cite{wang2021uformer} & \cite{mao2021deep} & -local\cite{chu2021revisiting}  & \textbf{ours} & \textbf{ours} \\
\hline
PSNR & 32.68 & 32.71 & 32.86 & 32.92 & 32.97 & 33.23 & 33.31 & 33.40 & 33.69 \\
SSIM & 0.959 & 0.959 & 0.961 & 0.961 & 0.967 & 0.963 & 0.964 & 0.965 & 0.967 \\
\hline
MACs(G) & 1235 & 170.7 & 169.5 & 140 & 89.5 & 187 & 778.2 & 84 & 65 \\
\hline
\end{tabular}
\end{center}

\end{table}
\begin{figure*}[!t]
\includegraphics[width=1.0\textwidth]{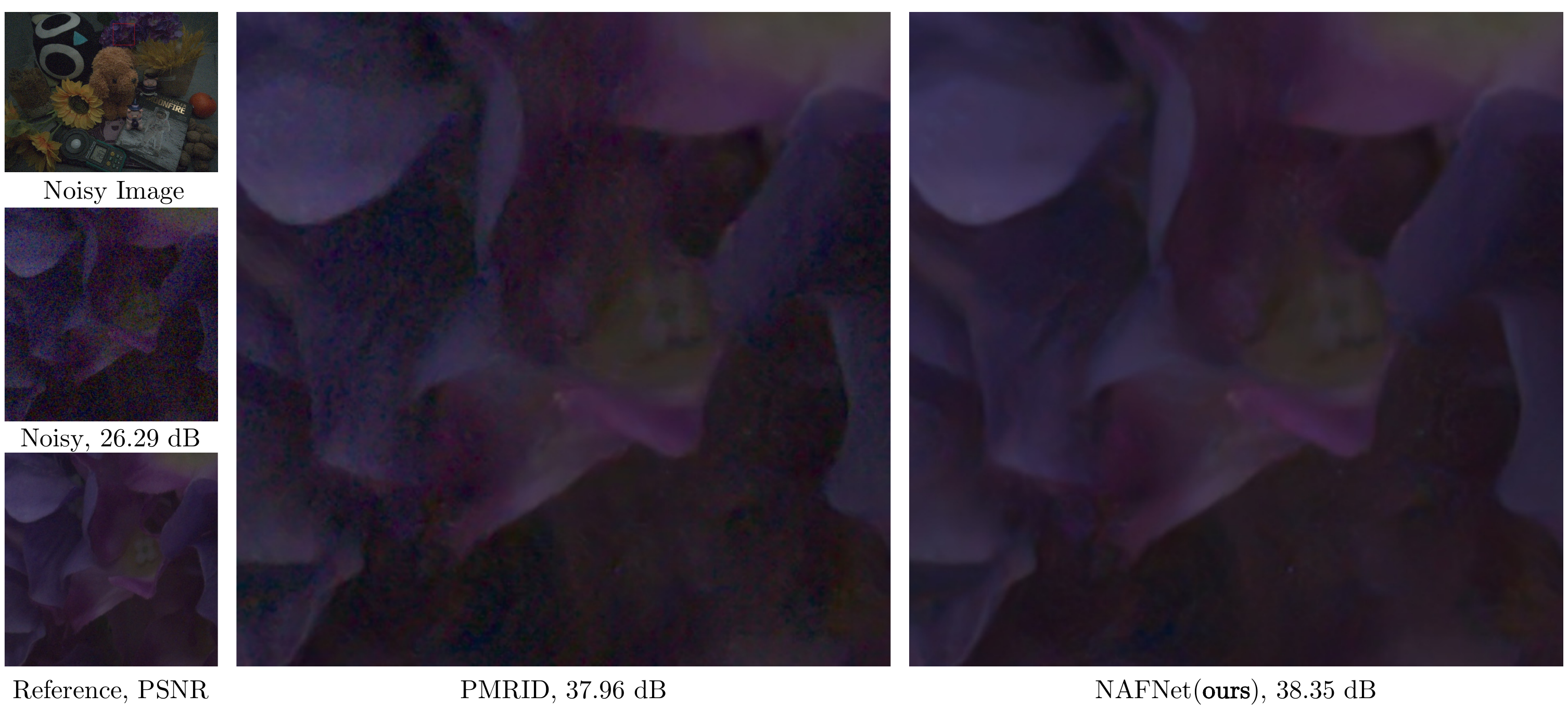}
\caption{Qualitatively compare the noise reduction effects of PMRID\cite{wang2020practical} and our porposed NAFNet. Zoom in to see details}
\label{fig:Raw-Denoise-Visual}
\end{figure*}
\begin{table}[ht]
\RawFloats
\scriptsize
\setlength{\tabcolsep}{3pt}
    \parbox{.48\linewidth}{
    \centering
\caption{Raw image denoising results on 4Scenes\cite{wang2020practical}
}
 \label{tab:SOTA.raw-denoise}
    \begin{tabular}{lccc}
{Method} & {PSNR}  & {SSIM} &{MACs(G)}  \\ 
\hline
PMRID\cite{wang2020practical} & 39.76 & 0.975 & 1.2  \\
{NAFNet(\textbf{ours})}  & 40.05&{0.977} & {1.1}\\ 
\hline
& & & \\
& & & 
\end{tabular}}
\hfill    
\parbox{.48\linewidth}{
\centering
\caption{Image deblurring results on REDS-val-300\cite{nah2021ntire}
}
 \label{tab:SOTA.REDS-val-300}

\begin{tabular}{lccc}
{Method} & {PSNR}  & {SSIM} & MACs(G)\\
\hline
 MPRNet\cite{waqas2021multi} & 28.79 & 0.811 & 776.7 \\
 HINet\cite{chen2021hinet} & 28.83 & 0.862 & 170.7 \\
 MAXIM\cite{tu2022maxim} & 28.93 & 0.865 & 169.5\\
 \hline
{NAFNet(\textbf{ours})}  & 29.09 & 0.867 & 65 \\
\hline      

\end{tabular}
}
\end{table}

\subsubsection{Image Deblurring} We compare the deblurring results of SOTA methods on GoPro\cite{nah2017deep} dataset, flip and rotate augmentations are adopted. As we shown in Table~\ref{tab:SOTAs-on-GoPro} and Figure~\ref{fig:PSNR_vs_MACs}, our baseline and NAFNet surpass the previous best method MPRNet-local\cite{chu2021revisiting} 0.09 dB and 0.38 dB in PSNR, respectively, with only 8.4\% of its computational costs. The visualization results are shown in Figure ~\ref{fig:deblur}, our baselines can restore sharper results compares to other methods. 
\subsubsection{Raw Image Denoising} We apply NAFNet to a raw image denoising task. The training and testing settings follow PMRID\cite{wang2020practical}, and we noted the testing set as 4Scenes (as the dataset contains 39 raw images of 4 different scenes in various light conditions) for simplicity. In addition, we make fair comparison by changing the width and number of blocks of NAFNet from 32 to 16, 36 to 7, respectively, so that the computational cost is less than PMRID. The results shown in Table~\ref{tab:SOTA.raw-denoise} and Figure~\ref{fig:Raw-Denoise-Visual} demonstrate NAFNet can surpass PMRID quantitatively and qualitatively. In addition, this experiment indicates our NAFNet can be scaled flexibly (from 1.1 GMACs to 65 GMACs).
\subsubsection{Image Deblurring with JPEG artifacts} We conduct experiments on REDS\cite{nah2021ntire} dataset, the training setting follows \cite{chen2021hinet,tu2022maxim}, and we evaluate the result on 300 images from the validation set (noted as REDS-val-300) following \cite{chen2021hinet,tu2022maxim}. As shown in Table~\ref{tab:SOTA.REDS-val-300},  our method outperforms other competing methods, including the previous winning solution (HINet) on the REDS dataset of NTIRE 2021 Image Deblurring Challenge Track2 JPEG artifacts\cite{nah2021ntire}.

\section{Conclusions}
By decomposing the SOTA methods, we extract the essential components and adopt them on a naive PlainNet. The obtained baseline reaches SOTA performance on image denoising and image deblurring tasks. By analyzing the baseline, we reveal that it can be further simplified: The nonlinear activation functions in it can be completely replaced or removed. From this, we propose a nonlinear activation free network, NAFNet. Although simplified, its performance is equal to or better than baseline. Our proposed baselines may facilitate the researchers to evaluate their ideas. In addition, this work has the potential to influence future computer vision model design, as we demonstrate that nonlinear activation functions are not necessary to achieve SOTA performance.
\subsubsection{Acknowledgements:} This research was supported by National Key R\&D Program of China (No. 2017YFA0700800) and Beijing Academy of Artificial Intelligence (BAAI). 

\appendix
\title{Appendix}
\author{}
\institute{}
\maketitle

\section{Other Details}

\subsection{Inverted Bottleneck}
Following ~\cite{liu2022convnet} we adopt inverted bottleneck design in the baseline and NAFNet. We first discuss the setting of the ablation studies.
In the baseline, 
the channel width within the first skip connection is always consistent with the input, its computational cost could be approximated by:
\begin{equation}\label{baseline-first}
H\times W \times c\times c + H\times W \times c \times k \times k + H\times W \times c \times c,
\end{equation}
where $H,W$ represent the spatial size of the feature map, $c$ indicates the input dimension, and $k$ is the kernel size of the depthwise convolution (3 in our experiments). In practice, $c \gg k\times k$, thus Eqn.~(\ref{baseline-first}) $\approx 2\times H\times W\times c\times c$.
The hidden dimension within the second skip connection is twice the input dimension, its computational cost is:
\begin{equation}\label{baseline-second}
H\times W \times c \times 2c + H\times W \times 2c \times c,
\end{equation}
notations following Eqn.~(\ref{baseline-first}). As a result, the overall computational cost of one baseline block $\approx 6\times H\times W\times c \times c$. 

As for NAFNet's block, the SimpleGate module shrinks the channel width by half. We double the hidden dimension in the first skip connection, and its computational cost could be approximated by:
\begin{equation}\label{naf-first}
H\times W \times c \times 2c + H \times W \times 2c \times k \times k + H\times W \times c \times c,
\end{equation}
notations following Eqn.~(\ref{baseline-first}). And the hidden dimension in the second skip connection follows baseline. Its computational cost is:
\begin{equation}\label{naf-second}
H\times W \times c \times 2c + H\times W \times c \times c.
\end{equation}
As a result, the overall computational cost of one NAFNet's block $\approx 6\times H \times W \times c \times c$, which is consistent with the baseline's block. The advantage of this is that the baseline and NAFNet can share hyperparameters, such as the number of blocks, learning rate, etc. 

As for the applications, the hidden dimension of the baseline's first skip connection is expanded to achieve better results. In addition, it should be noted that the above discussion omits the computation of some modules, e.g. layer normalization, GELU, channel attention, and etc., as their computational cost is negligible compared to convolution.

\subsection{Channel Attention and Simplified Channel Attention}
For a feature map with width of $c$,  the channel attention module shrinks it by a factor of $r$ and then project it back into $c$ (by fully-connect layer). The computational cost could be approximated by $c \times c/r + c/r \times c$. As to the simplified channel attention module, its computational cost is $c \times c$. For a fair comparison, we choose $r=2$ so that their computational costs are consistent in our experiments.

\subsection{Feature Fusion}
There are skip connections from the encoder block to the decoder block, and there are several ways to fuse the features of encoder/decoder. In \cite{chen2021hinet}, the encoder features are transformed by a convolution and then concatenate with the decoder features. In \cite{zamir2021restormer}, features are concatenated first and then transformed by a convolution. Differently, we simply element-wise add the encoder and decoder features as the feature fusion approach.

\begin{figure*}[!t]
\includegraphics[width=1.0\textwidth]{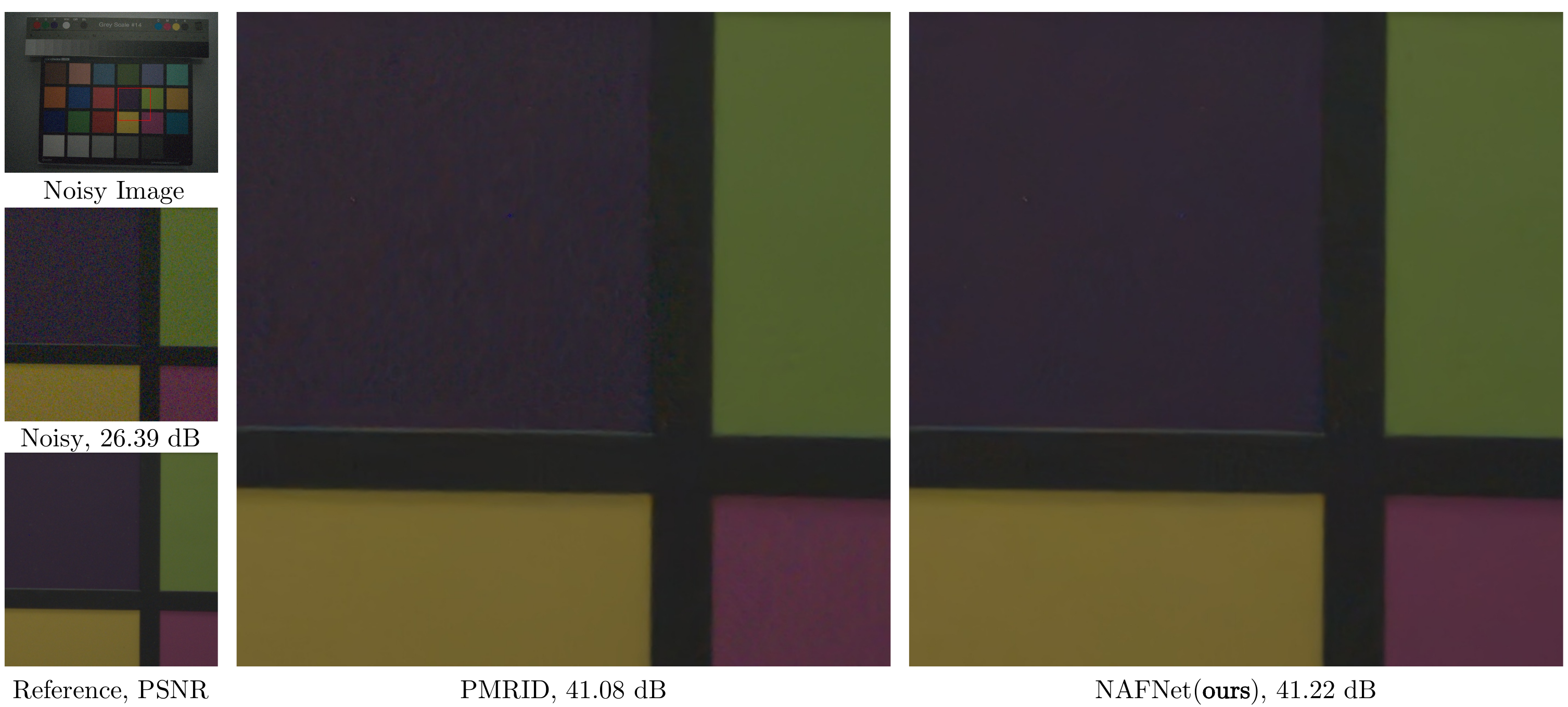}
\caption{Additional qualitatively comparison of raw image denoising results with PMRID\cite{wang2020practical}. Zoom in to see details}
\label{fig:more-Raw-Denoise-Visual}
\end{figure*}

\subsection{Downsample/Upsample Layer}
For the downsample layer, we use the convolution with a kernel size of 2 and a stride of 2. This design choice is inspired by \cite{alsallakh2020mind}. 
For the upsample layer, we double the channel width by a pointwise convolution first, and then follows a pixel shuffle module\cite{shi2016real}.



\section{More Visualization Results}
\begin{figure*}[!t]

\centering
\scalebox{0.98}{

\begin{tabular}{cc }
\includegraphics[width=0.48\textwidth]{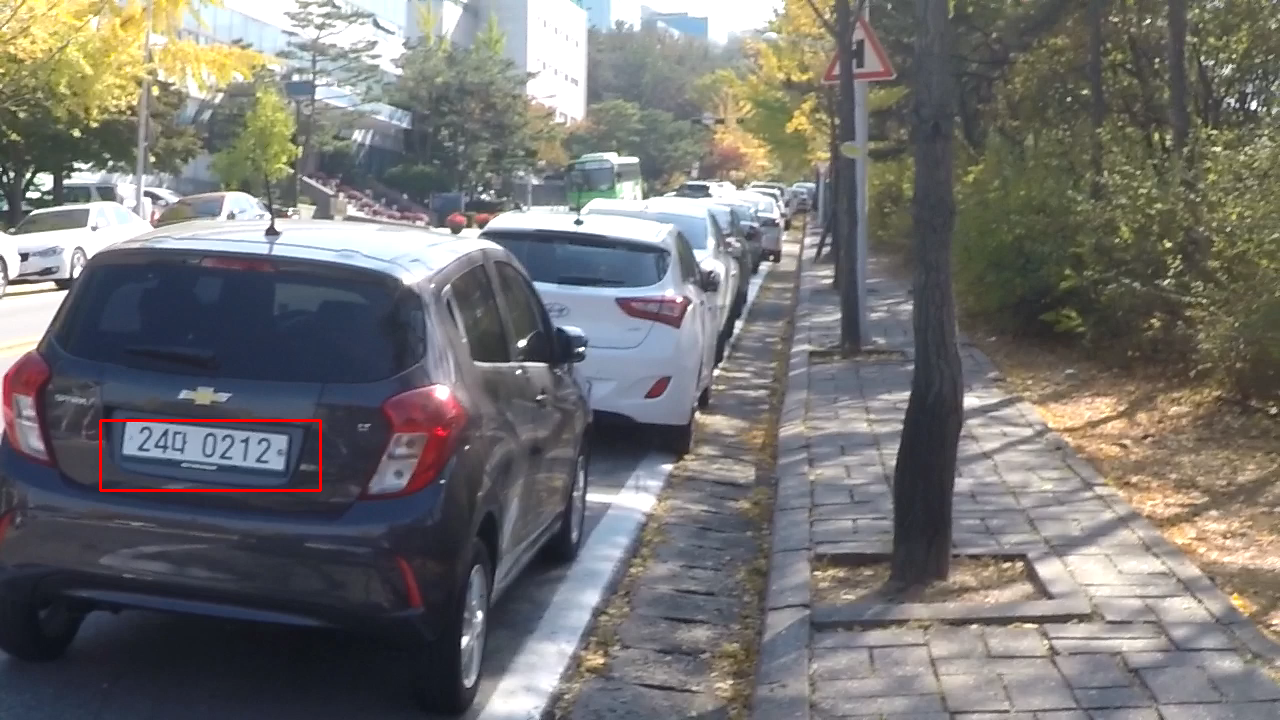} &
\includegraphics[width=0.48\textwidth]{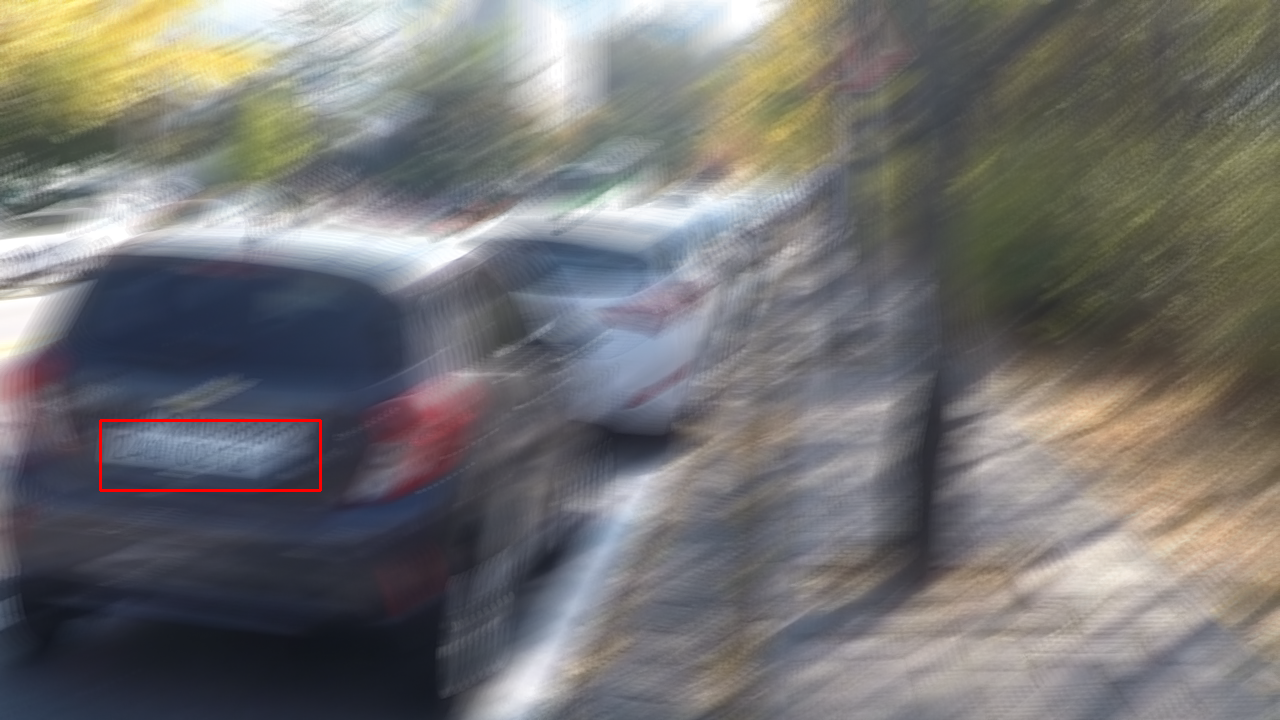} \\
PSNR & 19.45 dB \\
Reference & Blurry \\
\includegraphics[width=0.48\textwidth]{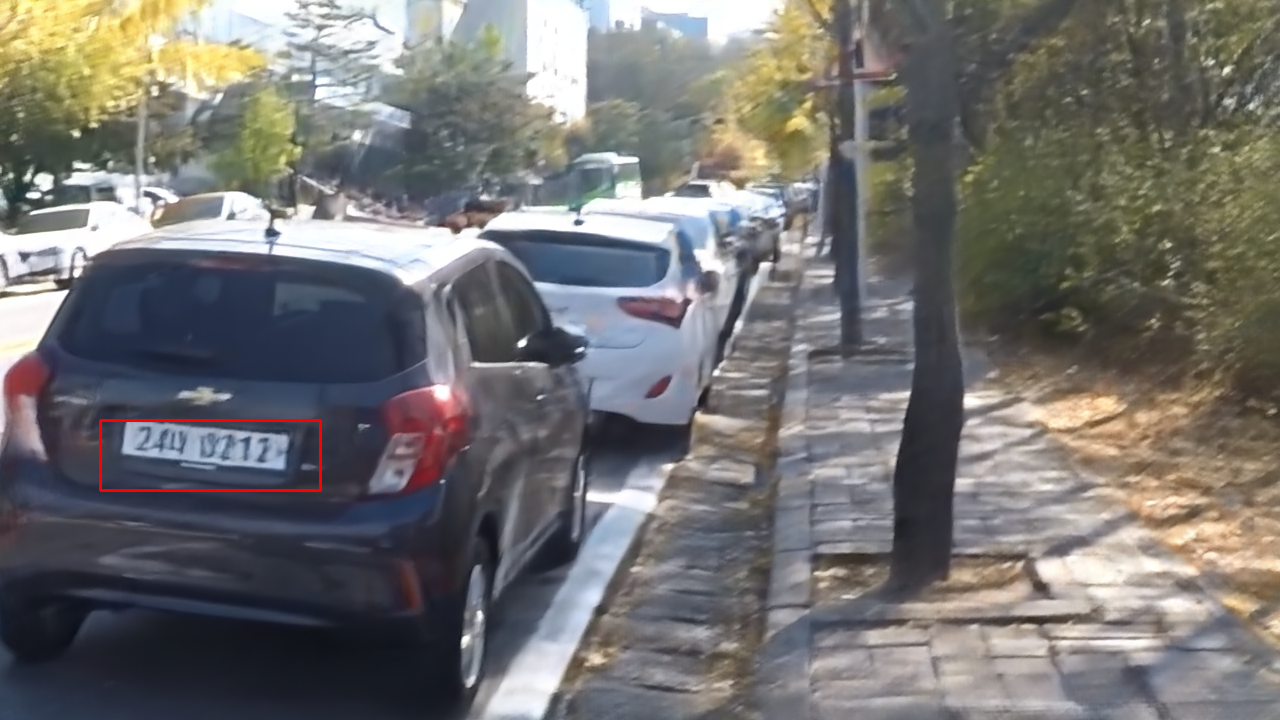} &
\includegraphics[width=0.48\textwidth]{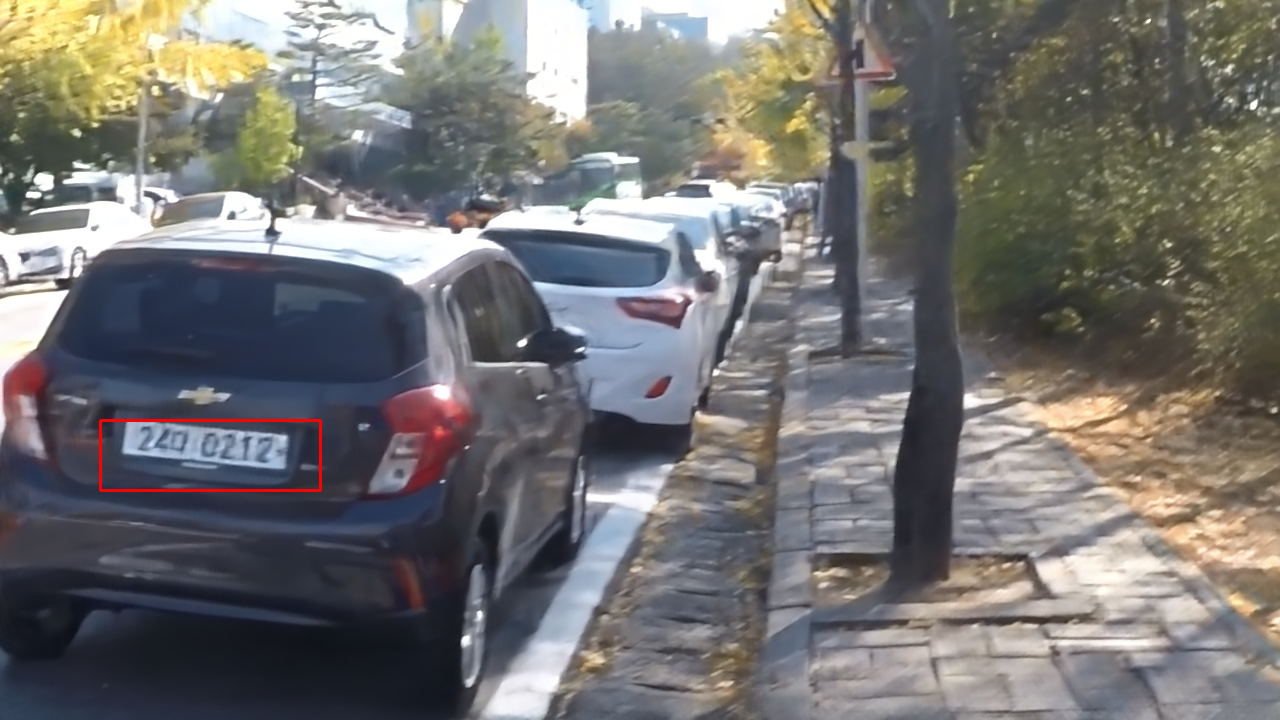} \\
25.81 dB & 27.03 dB \\
HINet\cite{chen2021hinet} & MPRNet-local\cite{chu2021revisiting}\\
\includegraphics[width=0.48\textwidth]{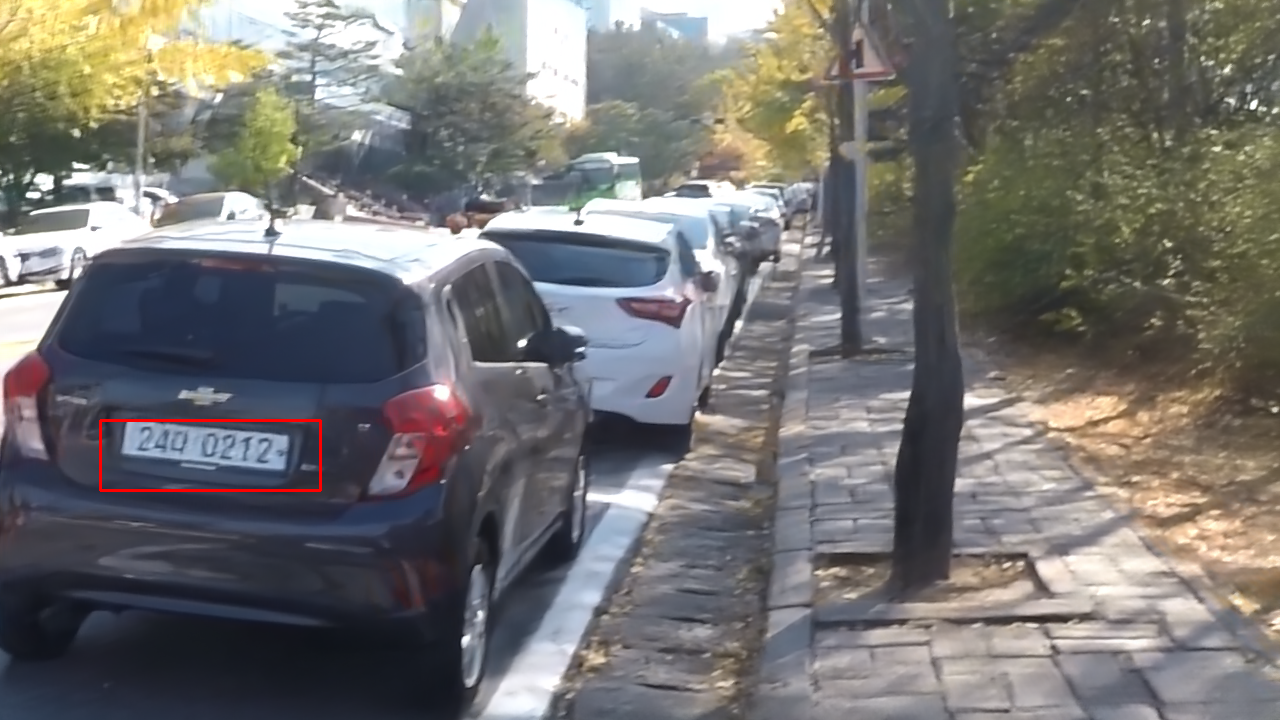} &
\includegraphics[width=0.48\textwidth]{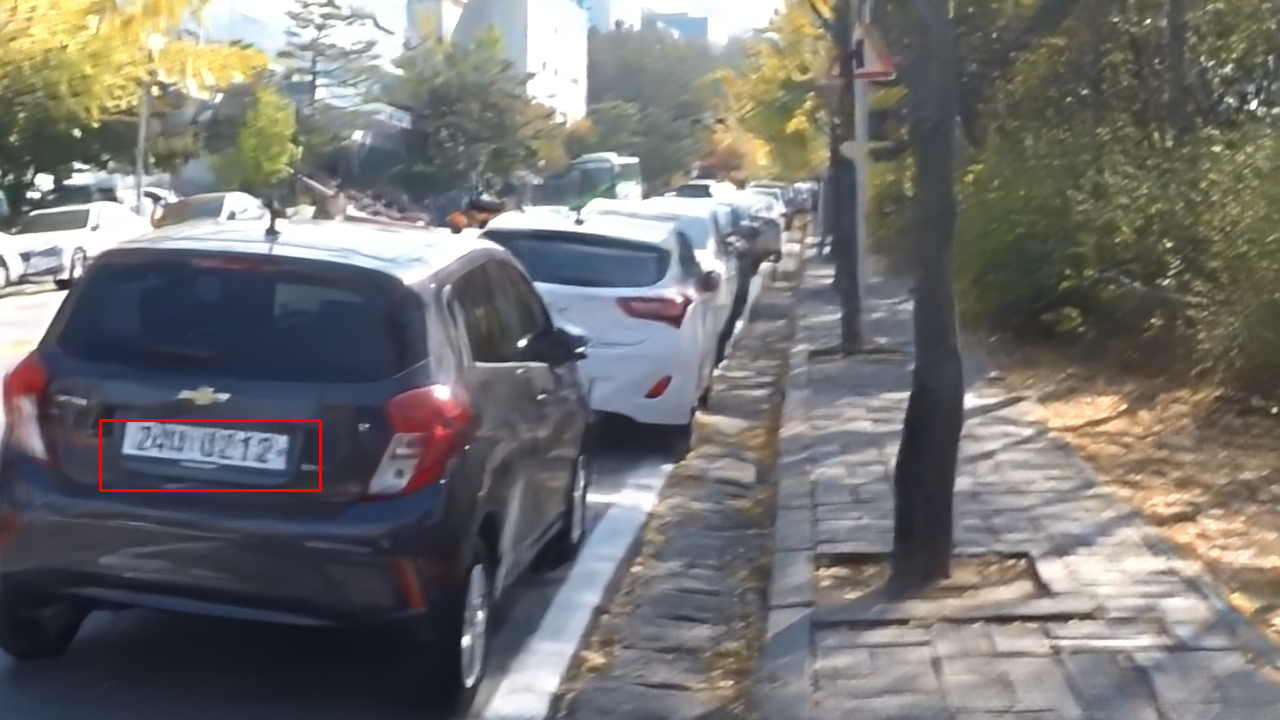} \\
26.96 dB & 25.67 dB \\
Restormer\cite{zamir2021restormer} & MPRNet\cite{waqas2021multi}\\
\includegraphics[width=0.48\textwidth]{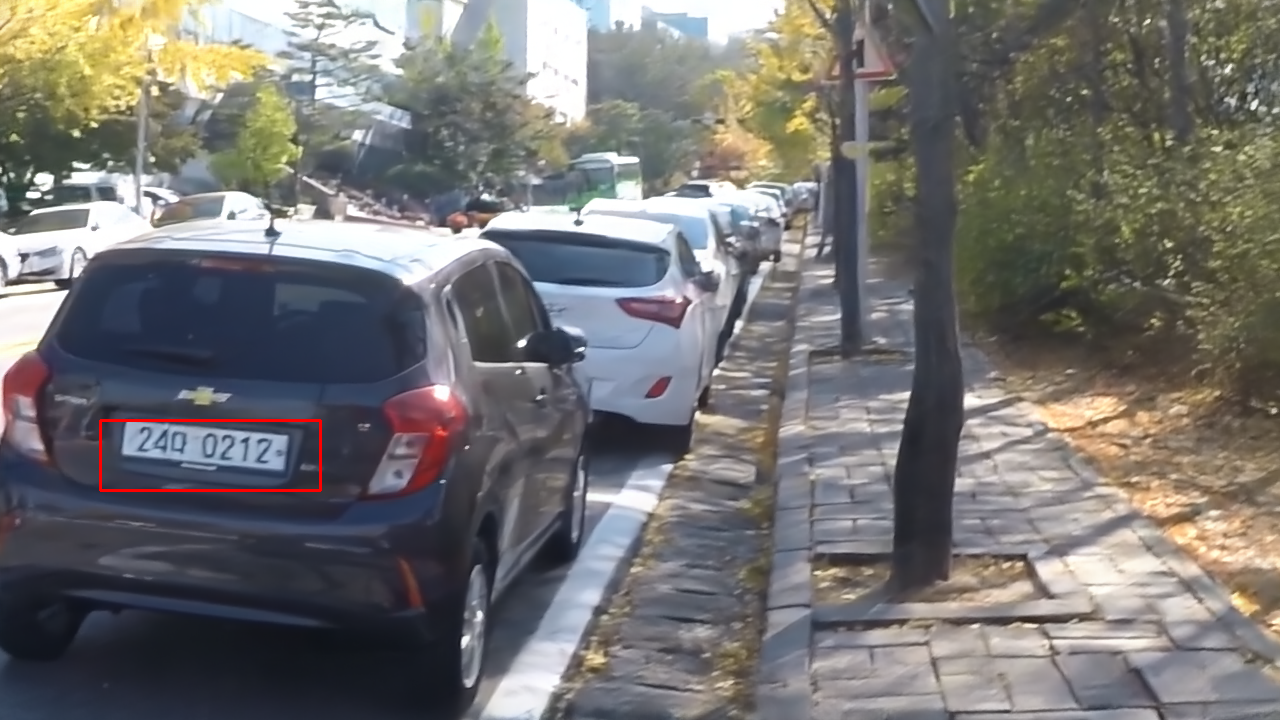} &
\includegraphics[width=0.48\textwidth]{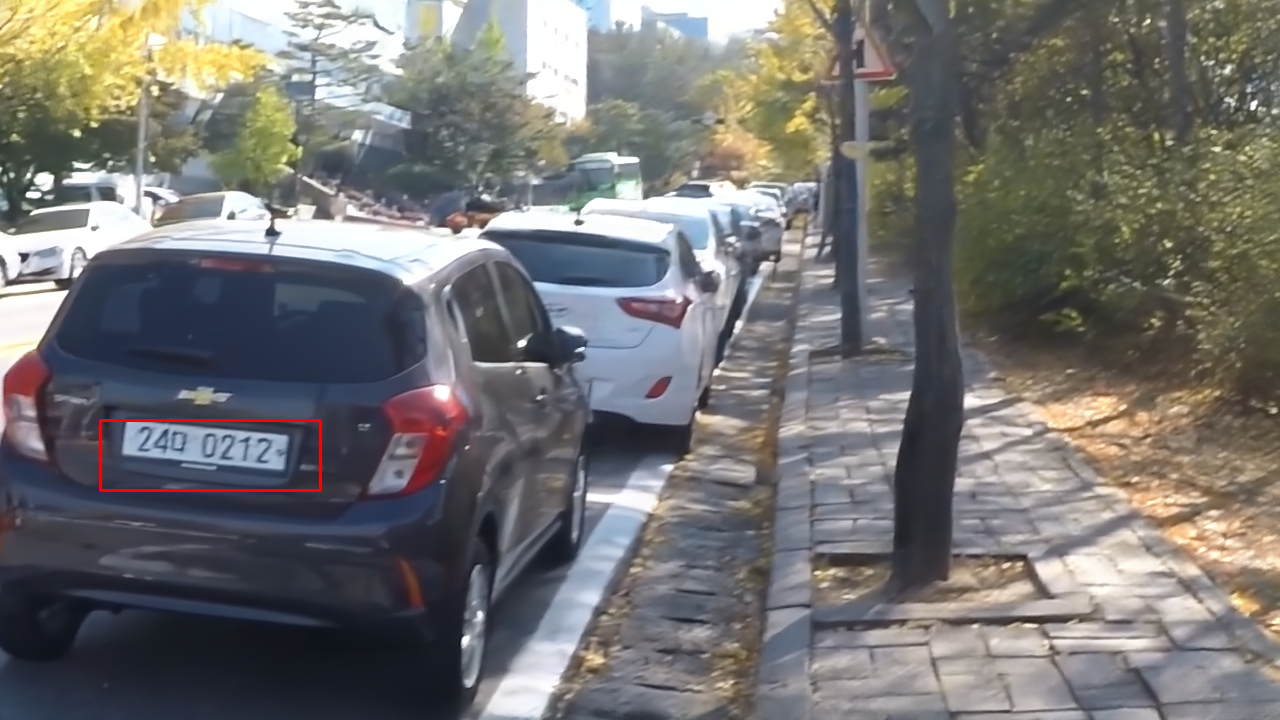} \\
28.11 dB & 28.71 dB \\
Baseline(\textbf{ours}) & NAFNet(\textbf{ours})\\
\end{tabular}

}
\caption{Additional qualitative comparison of image deblurring methods
}
\label{fig:more-visual-gopro1}
\end{figure*}

\begin{figure*}[!t]

\centering
\scalebox{0.98}{

\begin{tabular}{ccc }
\includegraphics[width=0.32\textwidth]{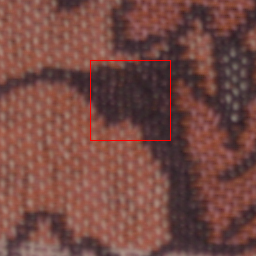} &
\includegraphics[width=0.32\textwidth]{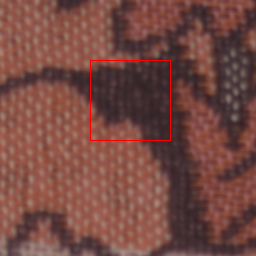} &
\includegraphics[width=0.32\textwidth]{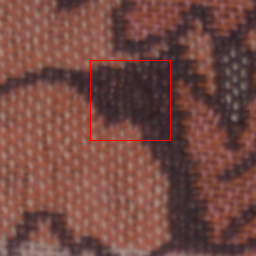} \\
PSNR & 38.89 dB & 41.35 dB \\
Reference & Restormer\cite{zamir2021restormer} & Baseline(\textbf{ours}) \\
\includegraphics[width=0.32\textwidth]{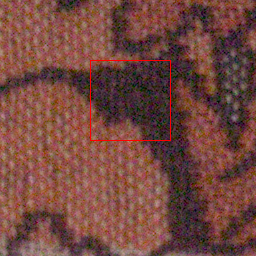} &
\includegraphics[width=0.32\textwidth]{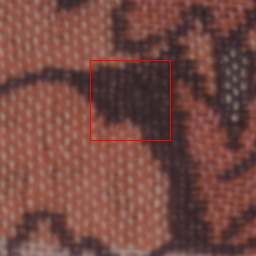} &
\includegraphics[width=0.32\textwidth]{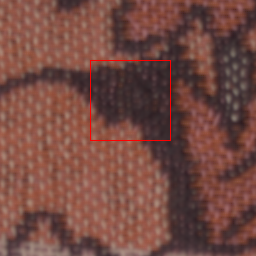} \\
24.99 dB& 38.41 dB& 41.45 dB\\
Noisy & MPRNet\cite{waqas2021multi}& NAFNet(\textbf{ours})\\

\includegraphics[width=0.32\textwidth]{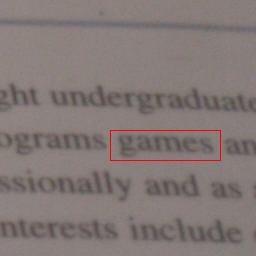} &
\includegraphics[width=0.32\textwidth]{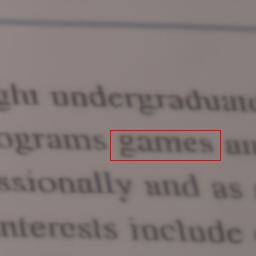} &
\includegraphics[width=0.32\textwidth]{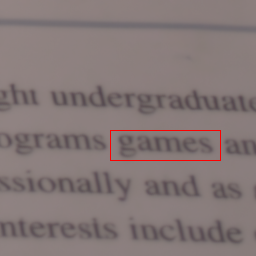} \\
PSNR & 37.91 dB & 38.25 dB \\
Reference & Restormer\cite{zamir2021restormer} & Baseline(\textbf{ours}) \\
\includegraphics[width=0.32\textwidth]{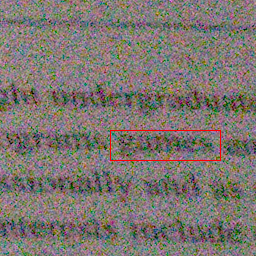} &
\includegraphics[width=0.32\textwidth]{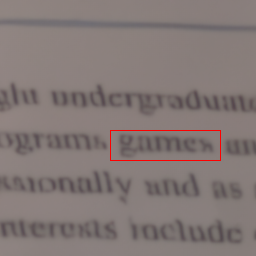} &
\includegraphics[width=0.32\textwidth]{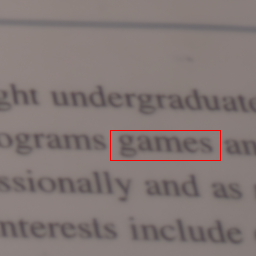} \\
19.05 dB& 37.29 dB& 38.74 dB\\
Noisy & MPRNet\cite{waqas2021multi}& NAFNet(\textbf{ours})\\


\end{tabular}

}
\caption{Additional qualitative comparison of image denoising methods
}
\label{fig:more-visual-sidd}
\end{figure*}

We provide additional visualization results of raw image denoising, image deblurring, RGB image denoising tasks, as we shown in Figure~\ref{fig:more-Raw-Denoise-Visual},~\ref{fig:more-visual-gopro1}, and ~\ref{fig:more-visual-sidd}. Our baselines can restore more fine details compare to other methods. It is recommended to zoom in to compare the details in the red box.

\clearpage
%
%
\bibliographystyle{splncs04}
\bibliography{egbib}

\end{document}